\documentclass[letterpaper, 10 pt, conference]{ieeeconf}  

\IEEEoverridecommandlockouts                              
\usepackage[utf8]{inputenc}
\usepackage[T1]{fontenc}

\usepackage{colortbl}

\usepackage[table,xcdraw]{xcolor}
\usepackage{amssymb}

\definecolor{rblue}{rgb}{0,0.5,1}
\definecolor{awesome}{rgb}{1.0, 0.13, 0.32}
\definecolor{hollywoodcerise}{rgb}{0.96, 0.0, 0.63}
\definecolor{lasallegreen}{rgb}{0.03, 0.47, 0.19}
\definecolor{hanpurple}{rgb}{0.32, 0.09, 0.98}
\definecolor{green(pigment)}{rgb}{0.0, 0.65, 0.31}

\definecolor{mygray}{gray}{.9}
\definecolor{mygreen}{RGB}{93,174,86}

\makeatletter
\let\NAT@parse\undefined
\makeatother
\usepackage[pagebackref=false,breaklinks=true,colorlinks,bookmarks=false]{hyperref}
\hypersetup{colorlinks=true,linkcolor={red},citecolor={hanpurple},urlcolor={magenta}}

\title{\LARGE \bf
EgoEvGesture: Gesture Recognition Based on Egocentric Event Camera
}

\author{Luming Wang$^{1}$, Hao Shi$^{1,*}$, Xiaoting Yin$^{1}$, Kailun Yang$^{2,3}$, Kaiwei Wang$^{1,*}$, and Jian Bai$^{1}$
\thanks{This work was supported by the National Natural Science Foundation of China (Grant No. 12174341 and No. 62473139) and Zhejiang Provincial Natural Science Foundation of China (Grant No. LZ24F050003), in part by the Hunan Provincial Research and Development Project (Grant No. 2025QK3019), in part by the Open Research Project of the State Key Laboratory of Industrial Control Technology, China (Grant No. ICT2025B20), and in part by Shanghai SUPREMIND Technology Company Ltd.}%
\thanks{$^{1}$The author is with the State Key Laboratory of Extreme Photonics and Instrumentation, Zhejiang University, China.}%
\thanks{$^{2}$The authors are with the School of Artificial Intelligence and Robotics, Hunan University, China.}%
\thanks{$^{3}$The authors are with the National Engineering Research Center of Robot Visual Perception and Control Technology, Hunan University, China.}%
\thanks{$^{*}$Corresponding authors.}
}

\usepackage{array}
\usepackage{booktabs}
\usepackage{graphicx}
\usepackage{cite}     
\usepackage{textcomp}
\usepackage{multirow}
\usepackage{comment}
\usepackage{amssymb}
\usepackage{amsmath}
\usepackage{subcaption}
\usepackage{pifont} 
\usepackage{rotating}
\usepackage{booktabs}    

\begin{document}

\maketitle
\thispagestyle{empty}
\pagestyle{empty}

\begin{abstract}

Egocentric gesture recognition is a pivotal technology for enhancing natural human-computer interaction, yet traditional RGB-based solutions suffer from motion blur and illumination variations in dynamic scenarios. While event cameras show distinct advantages in handling high dynamic range with ultra-low power consumption, existing RGB-based architectures face inherent limitations in processing asynchronous event streams due to their synchronous frame-based nature. Moreover, from an egocentric perspective, event cameras record data that includes events generated by both head movements and hand gestures, thereby increasing the complexity of gesture recognition. To address this, we propose a novel network architecture specifically designed for event data processing, incorporating (1) a lightweight CNN with asymmetric depthwise convolutions to reduce parameters while preserving spatiotemporal features, (2) a plug-and-play state-space model as context block that decouples head movement noise from gesture dynamics, and (3) a parameter-free Bins-Temporal Shift Module (BTSM) that shifts features along bins and temporal dimensions to fuse sparse events efficiently. We further establish the EgoEvGesture dataset, the first large-scale dataset for egocentric gesture recognition using event cameras. Experimental results demonstrate that our method achieves 62.7\% accuracy tested on unseen subjects with only 7M parameters, 3.1\% higher than state-of-the-art approaches. Notable misclassifications in freestyle motions stem from high inter-personal variability and unseen test patterns differing from training data. Moreover, our approach achieved a remarkable accuracy of 97.0\% on the DVS128 Gesture, demonstrating the effectiveness and generalization capability of our method on public datasets. The dataset and models are made available at \url{https://github.com/3190105222/EgoEv_Gesture}.

\end{abstract}

\section{Introduction}

\begin{figure}[!t]
  \centering
  \begin{subfigure}[b]{0.24\textwidth}
    \centering
    \includegraphics[width=\linewidth, height=0.1\textheight, keepaspectratio]{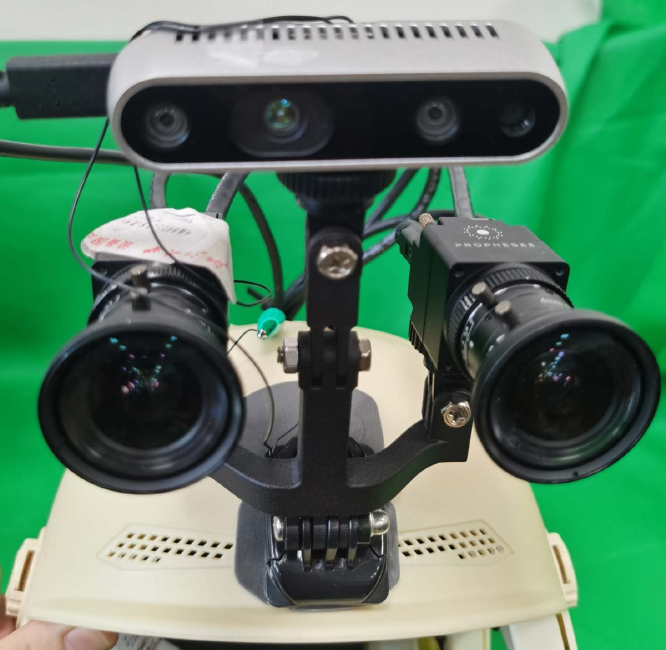}
    \caption{\centering Proposed head-mounted capture system.}
    \label{fig:setup:a}
  \end{subfigure}
  \hspace{-0.7em}
  \begin{subfigure}[b]{0.24\textwidth}
    \centering
    \includegraphics[width=\linewidth, height=0.1\textheight, keepaspectratio]{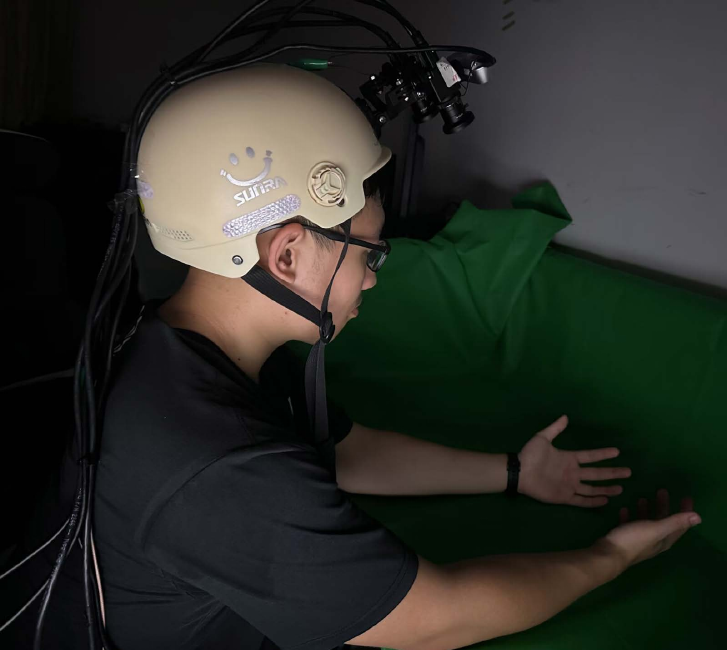}
    \caption{\centering Recording environment~(HDR).}
    \label{fig:setup:b}
  \end{subfigure}

  \begin{subfigure}[b]{\columnwidth}
    \centering
    \includegraphics[width=0.99\linewidth]{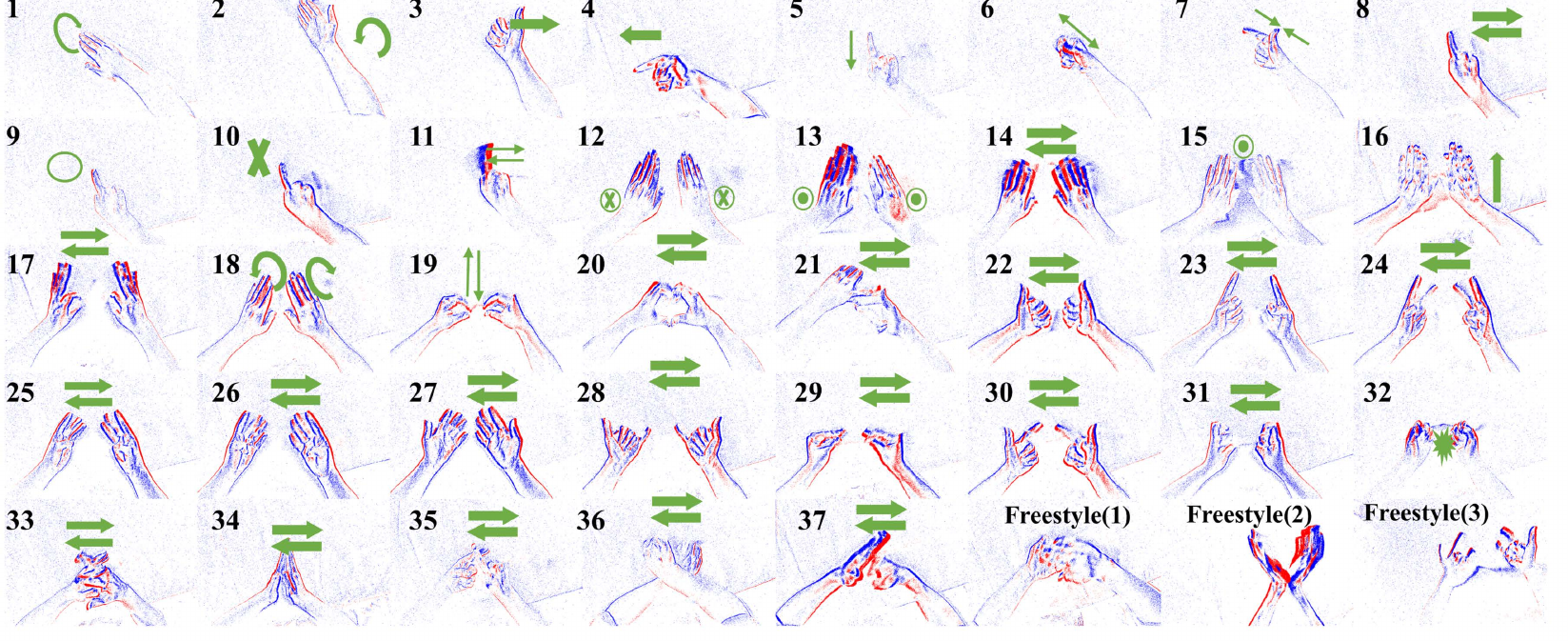}
    \caption{Representation of 38 gesture classes.}
    \label{fig:gestures}
  \end{subfigure}
  
  \caption{
    (a) Proposed head-mounted capture system (HMCS). It integrates two Prophesee EVK4 event cameras (left/right) and an Intel RealSense D435 RGB-D sensor (center).
    (b) Recording environment. Each participant is equipped with the HMCS and performs a series of 38 distinct gestures under normal light and low light conditions. 
    (c) The 38 classes of hand gesture designed in our established EgoEvGesture dataset. The figure shows events visualization, with arrows indicating the direction of hand (thick arrows) or finger (thin arrows) movements. The freestyle motions vary among participants, with three examples shown in the figure.
  }
  \vspace{-1.5em}
\end{figure}

Gesture recognition is a pivotal technology in fields such as computer vision~\cite{gao2024challenges}, human-computer interaction~\cite{woo2023survey}, virtual reality~\cite{bandini2020analysis}, and robotics~\cite{qi2024computer}. Egocentric gesture recognition, which captures actions from a first-person perspective, stands out for its natural and intuitive interaction modality, making it highly suitable for seamless integration into daily life. This has significant implications for Extended Reality (XR)~\cite{han2020megatrack}, smart wearable devices~\cite{zhang2018egogesture}, and human-computer interaction in various domains~\cite{mueller2017real} .
Unlike third-person settings, egocentric perspectives place sensors closer to the action, inherently incorporating natural attentional cues derived from the human line of sight. 

However, existing approaches that rely on traditional frame-based cameras are ill-suited for dynamic environments typical of egocentric perspectives. Traditional cameras, which capture images at fixed frame rates, often struggle with motion blur and sudden changes in lighting. These limitations are particularly pronounced in low-light conditions, where the reduced amount of available light exacerbates motion blur and makes it challenging to capture clear and usable images. Additionally, the continuous frame capture nature of traditional cameras results in high power consumption and memory requirements, further limiting their practicality for wearable devices and real-time applications~\cite{plizzari2024outlook, gallego2020event}.

Event cameras, on the other hand, offer a promising alternative. These cameras asynchronously measure intensity changes at the pixel level, generating neuromorphic events with very high temporal resolution (on the order of microseconds), virtually no motion blur, and high dynamic range (HDR, 120 dB). They also consume significantly less power and memory compared to traditional frame-based cameras, thus making them highly suitable for real-world applications such as egocentric action recognition on wearable devices~\cite{lan2023tracking,lin2024embodied,gallego2020event}. 
Despite these advantages, egocentric gesture recognition presents a significantly larger challenge for event cameras compared to conventional frame-based cameras: 1) In third-person settings, events are primarily generated by the motion of the observed subject. In contrast, in egocentric perspectives, events are generated by both the motion of the camera (due to head movements) and the motion of the hands performing gestures. This dual source of motion complicates the interpretation of event data, as it becomes difficult to disentangle the camera's ego-motion from the actual gesture movements. Consequently, the gap between third-person and egocentric gesture recognition is more pronounced for event cameras than for traditional cameras, making the task significantly more challenging. 
2) In addition, the development of egocentric gesture recognition using event cameras has been hindered by the lack of large-scale, well-annotated datasets as well.

To address these challenges, we first propose a novel approach with three progressive processing stages, as illustrated in Figure~\ref{arch}. First, we convert raw events into Locally Normalized Event Surfaces (LNES)~\cite{rudnev2021eventhands} that focus on the most recent temporal signature, suppressing historical ego-motion artifacts while maintaining real-time performance. 
Second, we employ Blaze~\cite{bazarevsky2019blazeface} and Vision Mamba (VMamba)~\cite{mambapy} blocks to extract spatiotemporal features.
The Blaze encoder-decoder, adapted from~\cite{bazarevsky2019blazeface} with proven efficiency in human-centric perception~\cite{millerdurai2024eventego3d}, processes each temporal bin efficiently. The VMamba block ~\cite{mambapy} addresses sparse event processing and long-term temporal modeling through discrete state space equations. 
Finally, we propose the Bins-Temporal Shift Module (BTSM), a zero-parameter mechanism that shifts features bidirectionally along two orthogonal axes:
1) Bins axis (fine temporal resolution: 33 ms/bin) for inter-bin context fusion,
2) Raw temporal axis (coarse resolution: 200 ms) for intra-bin motion alignment.
This dual-scale shifting resolves temporal fragmentation while maintaining motion invariance.

Due to the lack of event datasets in the egocentric setting, we developed EgoEvGesture, the first large-scale dataset for egocentric gesture recognition using event cameras. Captured under both normal and low-light conditions, it includes data from 10 subjects performing a variety of gestures, offering rich temporal and spatial information for training and evaluating models in real-world scenarios. Our proposed method achieves a classification accuracy of 62.7\% on this dataset, with only 7M parameters, outperforming existing (closely related) approaches by 3.0\%. 
To verify the effectiveness of our approach, we evaluated it on the DVS128 Gesture dataset~\cite{amir2017low}, achieving a competitive accuracy of 97.0\%. This result highlights the robustness of our architecture in handling sparse and temporally challenging event-based data. The significance of each proposed module is evaluated and confirmed in an ablation study.

In summary, this paper defines a new problem, i.e., gesture recognition based on an egocentric event camera, and makes the following technical contributions: 

\begin{itemize}
  \item \textbf{Construction of the first large-scale, real-world egocentric event-based gesture estimation dataset, EgoEvGesture}: Contains event sequences from 10 subjects performing 38 actions under both normal and low light conditions, with accurate action annotations.

  \item \textbf{Bins-Temporal Shift Module (BTSM)}: A zero-parameter mechanism enabling dual-axis feature shifting across bins and temporal dimensions to resolve motion fragmentation, achieving accuracy improvement without additional computational cost.
  
  \item \textbf{Novel neural architecture for egocentric event-based recognition}: The first end-to-end trainable framework specifically designed for gesture recognition from head-mounted event cameras, effective while maintaining computational efficiency.
\end{itemize}

\section{Related Work}

Next, we review related works on egocentric gesture recognition and event-based approaches to human perception. 
As shown in Table~\ref{tab:datasets}, our EgoEvGesture represents the first dataset for egocentric gesture recognition that uses real-world event streams, offering several advantages over existing RGB- or depth-based methods, as highlighted in the following sections.

\subsection{Egocentric Gesture Recognition}

Egocentric gesture recognition using RGB and depth sensors has been extensively explored~\cite{shamil2024utility,fan2024benchmarks,han2020megatrack,wen2023hierarchical}. The EgoGesture dataset~\cite{zhang2018egogesture}, one of the largest benchmarks, contains 83 gesture classes designed for wearable device interactions, which inspired our motion design. The EPIC-KITCHENS dataset~\cite{damen2020epic} focuses on action classification in kitchen environments, but prioritizes general activities over specific gestures. 
The FPHA dataset~\cite{garcia2018first} pioneered 3D hand pose annotations in egocentric scenarios but suffers from magnetic sensor interference and lacks two-hand pose data. Several datasets have advanced egocentric hand-object interaction research, yet each has limitations for dynamic gesture recognition. The H2O~\cite{kwon2021h2o} and Assembly101~\cite{sener2022assembly101} offer multi-view  videos but lacks gesture diversity, while ARCTIC~\cite{fan2023arctic} provides dexterous bimanual manipulation data but lacks dynamic gesture annotations. HOI4D~\cite{liu2022hoi4d} enriches single-hand interaction with 3D meshes and segmentation but omits two-hand scenarios.

Despite progress, RGB/depth-based methods remain constrained by motion blur, high power consumption, and limited dynamic capture capabilities, particularly in fast-moving or low-light conditions. These limitations have spurred the adoption of event cameras, which offer high temporal resolution and robustness to motion blur, opening new opportunities for gesture recognition.

\subsection{Event-based Methods for Human Perception}

\begin{table}[!t]
\centering
\caption{Dataset comparison with key attributes}
\label{tab:datasets}
\scriptsize
\setlength{\tabcolsep}{1pt}
\renewcommand{\arraystretch}{0.75}

\begin{tabular}{@{}>{\raggedright\arraybackslash}m{2cm}
                >{\centering\arraybackslash}m{0.7cm}
                >{\centering\arraybackslash}m{0.6cm}
                >{\centering\arraybackslash}m{0.5cm}
                >{\centering\arraybackslash}m{0.5cm}
                >{\centering\arraybackslash}m{0.6cm}
                >{\centering\arraybackslash}m{0.5cm}
                >{\centering\arraybackslash}m{0.5cm}
                >{\centering\arraybackslash}m{0.5cm}@{}}
\toprule
\bfseries Dataset & 
\rotatebox{55}{\bfseries Modality} & 
\rotatebox{55}{\bfseries Volume} & 
\rotatebox{55}{\bfseries Classes} & 
\rotatebox{55}{\bfseries 2H} & 
\rotatebox{55}{\bfseries Res} & 
\rotatebox{55}{\bfseries Year} & 
\rotatebox{55}{\bfseries Real} & 
\rotatebox{55}{\bfseries 1P} \\
\midrule

Ego-Gesture~\cite{zhang2018egogesture} & R & 3M & 83 & $\checkmark$ & 640 & 18 & $\checkmark$ & $\checkmark$ \\
FPHA~\cite{garcia2018first} & R & 0.1M & 45 & -- & 1080 & 18 & $\checkmark$ & $\checkmark$ \\
EPIC-K~\cite{damen2020epic} & R & 11M & 149 & $\checkmark$ & 1080 & 18 & $\checkmark$ & $\checkmark$ \\
H20~\cite{kwon2021h2o} & R & 0.6M & 36 & $\checkmark$ & 720 & 21 & $\checkmark$ & $\checkmark$ \\
Asm101~\cite{sener2022assembly101} & R & 110M & 1k & -- & 1080 & 21 & $\checkmark$ & $\checkmark$ \\
ARCTIC~\cite{fan2023arctic} & R & 2M & 11 & $\checkmark$ & 2.8k & 24 & $\checkmark$ & $\checkmark$ \\
HOI4D~\cite{liu2022hoi4d} & R & 2M & 54 & -- & 8k & 24 & $\checkmark$ & $\checkmark$ \\
DVS128~\cite{amir2017low} & E & 1kS & 10 & -- & 128 & 17 & $\checkmark$ & -- \\
N-EPIC~\cite{plizzari2022e2} & E & NA & 8 & $\checkmark$ & 225 & 22 & -- & $\checkmark$ \\
EHoA~\cite{chen2024ehoa} & E+R & 2kS & 8 & -- & 346 & 24 & $\checkmark$ & -- \\
EvRealH~\cite{jiang2024evhandpose} & E+R & 79m & NA & -- & 720 & 23 & $\checkmark$ & -- \\
Ev2H-S~\cite{millerdurai20243d} & E & * & NA & $\checkmark$ & 512 & 23 & -- & -- \\
Ev2H-R~\cite{millerdurai20243d} & E+R & 20m & NA & $\checkmark$ & 346 & 23 & $\checkmark$ & -- \\
EvHands~\cite{rudnev2021eventhands} & E & 100h & NA & -- & 240 & 21 & -- & -- \\
EvEgo3D~\cite{millerdurai2024eventego3d} & E & 127m & NA & -- & 320 & 24 & B & $\checkmark$ \\
Helios~\cite{bhattacharyya2024helios} & E & 600S & 7 & -- & 320 & 24 & $\checkmark$ & $\checkmark$ \\
EvRealH2~\cite{jiang2024complementing} & E+R & 74m & NA & -- & 346 & 24 & $\checkmark$ & -- \\
Ours & E & 5kS & 38 & $\checkmark$ & 720 & 25 & $\checkmark$ & $\checkmark$ \\
\bottomrule
\end{tabular}

\vspace{0.1em}
\parbox{\linewidth}{\scriptsize
$\checkmark$=Yes, --=No, B=Both. \\
\textbf{Attributes:} Modality (R=RGB, E=Event), 2H: Two-Hand, Res: Resolution (pixels), Real: Real-World, 1P: 1st-Person \\
\textbf{Units:} M=10\textsuperscript{6}, k=10\textsuperscript{3}, S=Samples, m=minutes, h=hours, *=3.1$\times$10\textsuperscript{8} events}

\vspace{-2em}
\end{table}

Event cameras, inspired by biological vision systems, asynchronously record pixel intensity changes, making them highly suitable for dynamic scenes with fast motion~\cite{gallego2020event}. Recent works have explored the use of event cameras for human perception tasks such as human pose estimation~\cite{millerdurai2024eventego3d}, action classification~\cite{amir2017low}, eye tracking~\cite{li2024gaze}, sign language recognition~\cite{zhang2024evsign}, and head pose estimation~\cite{yuan2024event}, demonstrating their advantages in temporal resolution and robustness to motion blur. Most existing event-based studies on hand motion analysis are conducted from a third-person perspective. DVS128~\cite{amir2017low} proposed the first event-based hardware system for gesture recognition. However, this system faces challenges such as low resolution and a lack of hand pose diversity. EventHands~\cite{rudnev2021eventhands}  proposes Locally Normalized Event Surfaces (LNES) to aggregate asynchronous event data into a compact 2D representation and achieves 1kHz 3D hand pose estimation. EvHandPose~\cite{jiang2024evhandpose} employs a novel hand flow representation and a weakly-supervised framework to address motion ambiguities in event streams, though its focus lies in reconstruction rather than gesture recognition. Ev2Hands~\cite{millerdurai20243d} achieved the first 3D tracking of two fast-moving and interacting hands from a single monocular event camera. EHoA~\cite{chen2024ehoa} pioneered hand-object action recognition via event vision. EvRGBHand~\cite{jiang2024complementing} introduced complementary use of event streams and RGB frames for hand mesh reconstruction. E2(GO)MOTION~\cite{plizzari2022e2} synthesizes the N-EPIC-Kitchens dataset, the first egocentric action recognition dataset based on event data.
The recent work Helios~\cite{bhattacharyya2024helios} proposes an event-based gesture recognition system optimized for smart eyewear, trained on a synthetic 7-class dataset to demonstrate low-power real-time operation. However, it shows a significant real-world accuracy drop, limited gesture diversity, and lacks dataset openness, while synthetic training introduces domain gap challenges.

The unique asynchronous and sparse nature of event data conflicts with conventional synchronous frame-based RGB processing. Despite microsecond temporal resolution, this ultra-high temporal granularity introduces unique computational challenges. Existing event-based approaches predominantly focus on instantaneous feature extraction tasks rather than continuous temporal understanding. This limitation stems from two inherent properties: (1) The asynchronous discrete nature of event representation complicates the modeling of coherent temporal dynamics, and (2) The fragmented temporal cues inherent in sparse event streams hinder process-level semantic interpretation. Consequently, current methodologies struggle to reconcile the conflict between leveraging fine-grained motion dynamics captured by event sensing and establishing holistic temporal reasoning frameworks. This modality discrepancy underscores the critical need for dedicated architectures addressing temporal continuity in event-based gesture recognition.

A comprehensive comparison of existing gesture recognition datasets across modalities is presented in Table~\ref{tab:datasets}. The table systematically evaluates 18 datasets through eight critical dimensions: sensing modality, data volume, action classes, two-hand interaction support, spatial resolution, publication year, real-world applicability, and first-person perspective. Notably, EgoEvGesture (our proposed dataset) establishes four key advantages: (1) First and only event-based dataset for egocentric gestures – Existing event datasets focus on third-person actions, while ours captures head-motion-contaminated events unique to first-person view. (2) the largest event-based sample volume (5,419 samples), (3) the richest gesture vocabulary among event-based datasets (38 classes), and (4) native support for two-hand interaction in egocentric scenarios - features previously unavailable in existing event-stream benchmarks.

\begin{figure}
  \centering
  \includegraphics[width=0.98\linewidth,height=0.15\textheight]{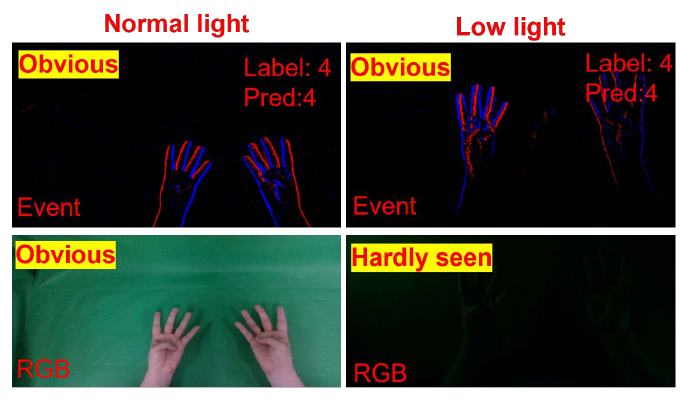}
  \vspace{-0.5em}
  \caption{A comparison of what an event camera (top) and an RGB camera (bottom) captured under normal light (left) and low light (right) conditions. Event cameras are almost unaffected by lighting conditions, while RGB images are barely distinguishable in low-light conditions.}
  \vspace{-1.5em}
  \label{fig:lighting_comparison}

\end{figure}

\section{EgoEvGesture Dataset} 
\label{sec:dataset}
\subsection{Head-Mounted Capture System~(HMCS)} 
\label{subsec:hw_config}

\begin{figure}[!t]
    \centering
    \includegraphics[width=0.99\linewidth]{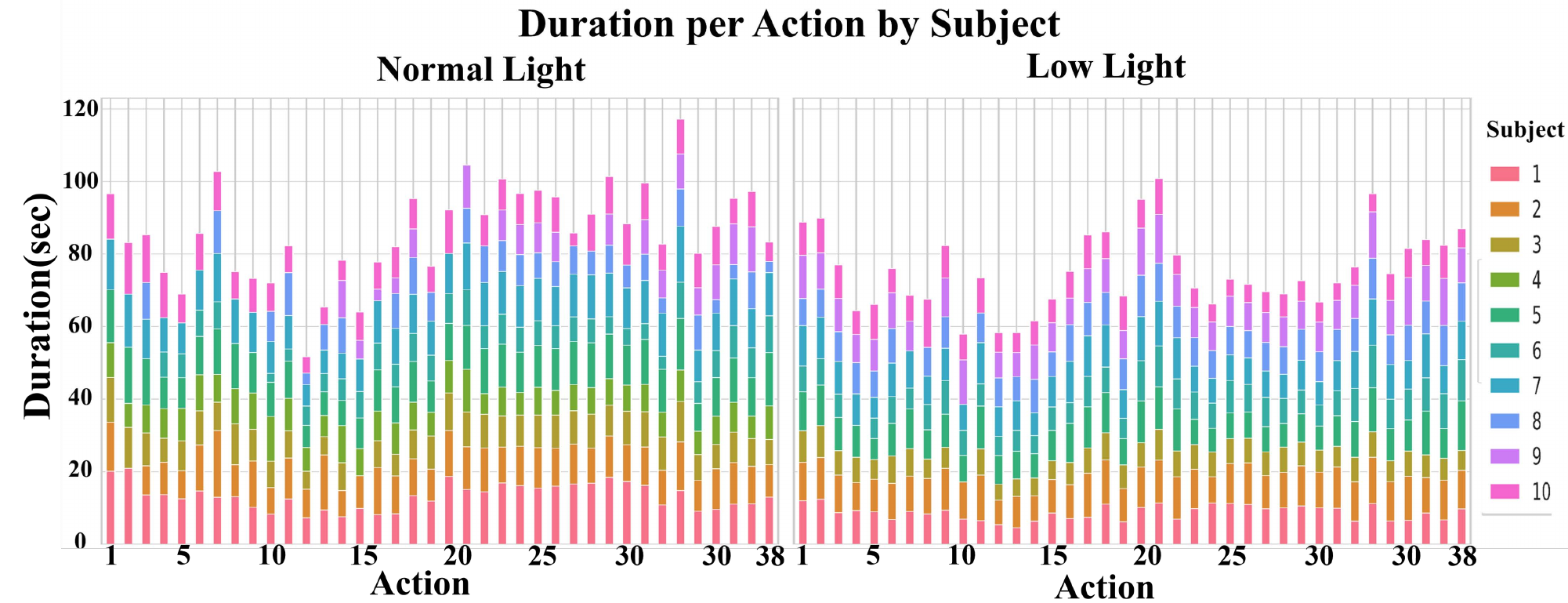}
    \caption{Action duration distributions across subjects under normal light (left) and low light (right) conditions. Each vertical bar represents the cumulative duration for a specific action, with colored segments indicating subject-wise contributions.}
    \label{fig:duration}
    \vspace{-1.5em}
\end{figure}

Our egocentric gesture acquisition platform (illustrated in Figure~\ref{fig:setup:a}) employs a custom-designed head-mounted capture system (HMCS) comprising a lightweight bicycle helmet modified with 3D-printed mounts to rigidly affix two Prophesee EVK4 event cameras and an Intel RealSense D435 depth camera. Temporal synchronization between event streams and RGB-D frames is achieved through hardware triggering, where the Intel RealSense D435 emits a synchronization signal to align data acquisition across all sensors. The event cameras utilize 4 mm focal length lenses (MV-LD-4-4M-G) providing a 94{\textdegree} horizontal field of view (FOV) with a native resolution of $1280 \times 720$ pixels. Each pixel has a size of $4.86\mu m \times 4.86\mu m$ with temporal resolution up to 1 MHz, capturing brightness changes through asynchronous events. For normal light conditions, we employed the default EVK4 threshold configuration (0\% bias\_diff\_off and 0\% bias\_diff\_on). In low-light scenarios, sensitivity was enhanced by configuring 20\% bias\_diff\_off and 20\% bias\_diff\_on parameters to increase event generation rate. The Intel RealSense D435 RGB-D camera operates at 30 fps with synchronized $1280 \times 720$ resolution for depth streams, which may be used for further research. The system connects via USB 3.0 to a host computer for power delivery and real-time data streaming, enabling uninterrupted recording sessions. The compact design and flexible HMCS allow users to freely move their heads and perform hand movements while ensuring precise temporal alignment.

\subsection{Data Collection Protocol}
\label{subsec:collection }

The dataset consists of data from 10 participants, encompassing a diverse range of body types and genders. Using the HMCS (Figure~\ref{fig:setup:a}), each participant performes 38 categorized gestures (Figure~\ref{fig:gestures}) under both normal and low-light conditions (Figure~\ref{fig:setup:b}), with at least four repetitions per gesture-condition combination, ensuring a robust and varied dataset. The gestures are categorized into three main groups: 

\begin{itemize}
\item \textbf{Single-Hand Gestures (Right Hand):} 11 classes including
Wave to the right, Wave to the left, Right thumb points to the right, Right index finger points to the left, Tap, Spread, Shrink, Swipe, Draw a circle, Draw an X, and NO (Shake finger).
\item \textbf{Two-Hand Gestures (No Occlusion):} 21 classes including Wave down, Wave up, Wave normally (arms close and move away), Raise arms, Push forward, OK, Flip palms, Grab, Make a heart, Camera gesture, Thumbs up, Show number 1-9, and Make a fist.
\item \textbf{Two-Hand Gestures (Mutual Occlusion):} 6 classes emphasizing challenging hand-hand interactions including Interlock fingers, Clap hands, Cross index fingers, Rotate arms, Salute with a fist, and Freestyle motion.
\end{itemize}

Figure~\ref{fig:lighting_comparison} comparatively analyzes RGB and event camera outputs under varying illumination conditions, demonstrating the latter's superior robustness to lighting variations.

\subsection{Dataset Composition and Statistics}
\label{subsec:statistics }
\begin{figure}[!t]
\centering

  \centering
  \includegraphics[width=0.99\linewidth, height=0.145\textheight]{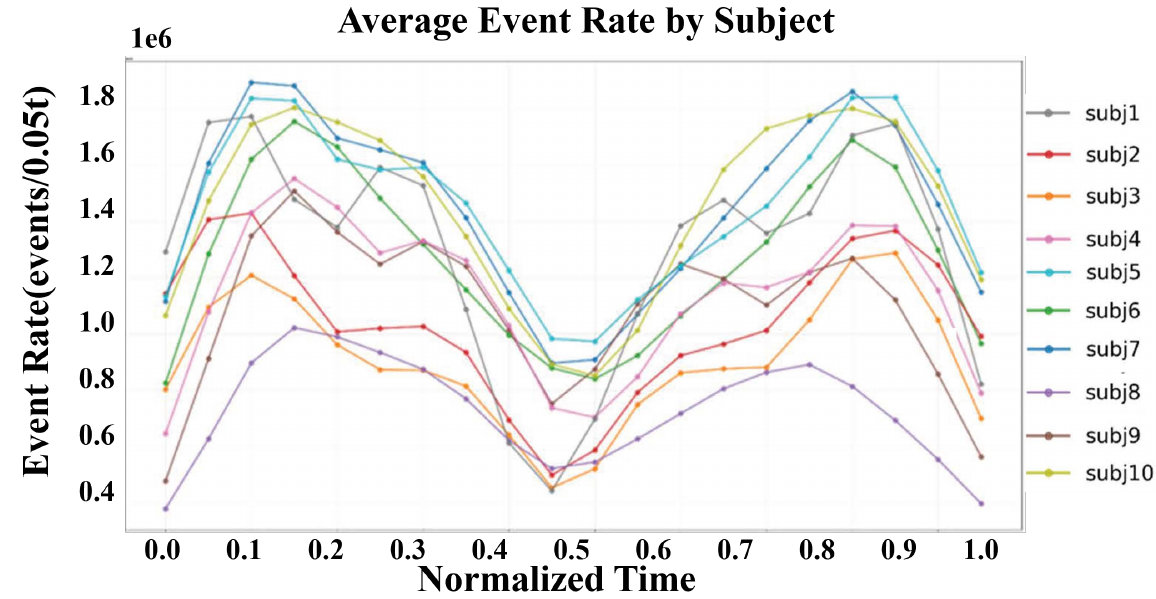}
  \vspace{-0.5em}
  \caption{Subject-wise normalized event rates showing individual adaptation variance.}
  \vspace{-2em}
  \label{fig:action_specificity}

\end{figure}

The dataset provides 5,419 finely segmented event data ($\sim$193 GB) with corresponding labels based on the precise start and end times of each action. The dataset exhibits significant variability, as evidenced by two key characteristics: (1) the duration distribution of gestures varies across participants, reflecting individualized articulation speeds (Figure~\ref{fig:duration}); and (2) substantial differences in normalized event rates are observed among different subjects (Figure~\ref{fig:action_specificity}). 
This variability in action execution patterns is crucial for training robust gesture recognition models to generalize across users and lighting. 
Event camera data, less lighting-sensitive than RGB, further enhances the dataset's utility for research in challenging environments.

In summary, the established EgoEvGesture dataset offers a rich and diverse collection of egocentric gesture data, captured under varying lighting conditions, making it a valuable resource for advancing research in gesture recognition using event cameras.

\section{Methodology}
\subsection{Architecture Overview: Dual-Motion Disentanglement Framework}
To address the critical challenge of ego-hand motion entanglement in event streams while capturing holistic temporal patterns, we propose a temporal pyramid architecture with three progressive processing stages (Figure~\ref{arch}):

\begin{figure*}[!t]
\centering
\includegraphics[width=0.95\linewidth, height=0.3\textheight, keepaspectratio]{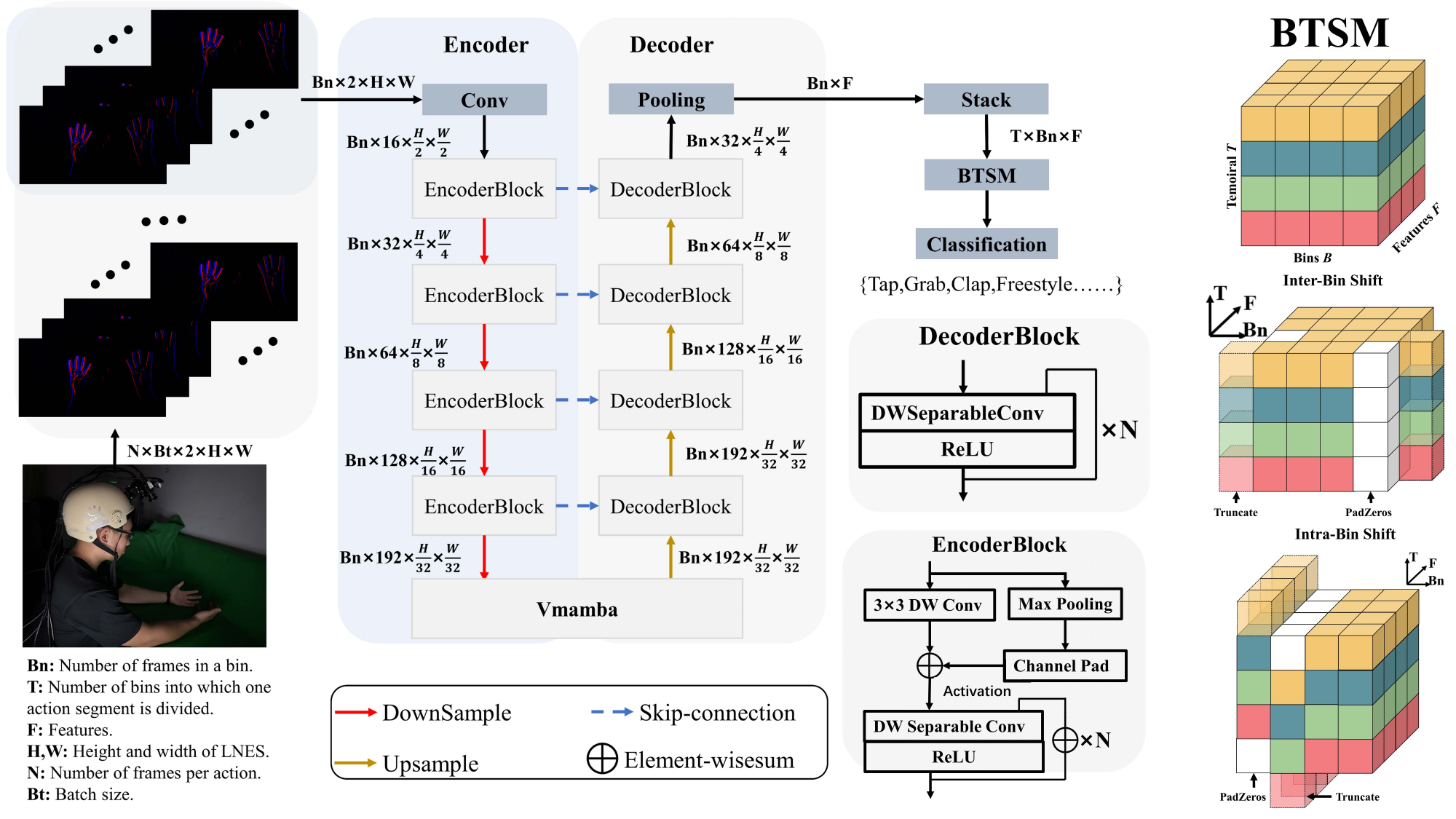}
\caption{Three-stage processing pipeline: (1) LNES converts raw events to normalized surfaces preserving recent temporal signatures; (2) Blaze and VMamba blocks extract multi-scale features; (3) BTSM aligns temporal context.}
\label{arch}
\vspace{-1.5em}
\end{figure*}

We first convert raw events into Locally Normalized Event Surfaces (LNES) that focus on the most recent temporal signature to suppress historical ego-motion artifacts while maintaining real-time performance. We then adopt Blaze \& VMamba blocks to extract spatiotemporal features. Our Blaze encoder-decoder, adapted from~\cite{bazarevsky2019blazeface} with proven efficiency in egocentric human perception~\cite{millerdurai2024eventego3d}, processes each temporal bin. The VMamba block addresses sparse event processing and long-term temporal modeling through discrete state space equations (detailed in Sec.~\ref{sec:vmamba}). Finally, we employ the Bins-Temporal Shift Module (BTSM) to resolve temporal semantic fragmentation caused by segmented processing and perform temporal alignment (detailed in Sec.~\ref{sec:btsm}).

\subsection{VMamba Block} \label{sec:vmamba}
Despite the ability to adapt to high dynamic ranges, event-based gesture recognition faces challenges due to the sparsity of events, which makes it difficult for traditional methods to effectively focus on key areas. Meanwhile, the very high event resolution, while enhancing the ability to capture fine details, also makes long-term temporal understanding more challenging. To resolve these, we introduce the VMamba block, which bridges the modality gap via adaptive computation allocation — dynamically prioritizing active temporal regions while preserving temporal fidelity. Specifically, the VMamba block employs density-aware gating to route computational resources toward high-event-density intervals and suppress inactive periods, ensuring efficient feature extraction without sacrificing microsecond-level temporal granularity.

The VMamba block offers three key advantages:
\begin{itemize}
\item \textbf{Plug-and-Play Compatibility}: Maintains original feature dimensions through dimension-preserving operations, enabling seamless integration into existing architectures.

\item \textbf{Sparse Event Adaptation}: Selectively focuses computation on active temporal regions through adaptive temporal attention, effectively handling sparse event patterns.

\item \textbf{Linear Complexity Scaling}: Achieves efficient computation with O($N^2 + NF$) complexity, where $N$ denotes the number of temporal steps and $F$ represents the feature dimension. The $NF$ term specifically captures the linear scaling of feature transformation costs across temporal windows, enabling practical processing of long videos.
\end{itemize}

\subsection{Bins-Temporal Shift Module (BTSM)}\label{sec:btsm}
Existing event-based approaches predominantly focus on instantaneous feature extraction tasks (e.g., 3D keypoint estimation) rather than continuous temporal understanding. This limitation arises from two key properties: (1) the asynchronous discrete nature of event representation complicates the modeling of coherent temporal dynamics, and (2) the fragmented temporal cues inherent in sparse event streams hinder process-level semantic interpretation. Consequently, current methodologies struggle to reconcile the conflict between leveraging fine-grained motion dynamics captured by event sensing and establishing holistic temporal reasoning frameworks. This modality discrepancy underscores the critical need for dedicated architectures addressing temporal continuity in event-based gesture recognition.
To tackle these challenges, the Bins-Temporal Shift Module (BTSM) implements dual-stage temporal modeling through cascaded shift operations including Intra-Bin Fusion and Inter-Bin Fusion. This parameter-free design enables comprehensive temporal modeling through layered feature propagation and addresses temporal semantic fragmentation caused by segmented processing.

\subsubsection{Dual-Phase Temporal Fusion}
The proposed temporal fusion comprises two complementary phases sharing identical feature transformation mechanics but operating along different axes:

\begin{equation}
\mathcal{T}(X, d) = \text{concat}\left(
\begin{aligned}
    &\text{roll}(X_{:,:,:,0:\frac{F}{4}}, \text{shift}=-1, \text{dim}=d), \\
    &\text{roll}(X_{:,:,:,\frac{F}{4}:\frac{F}{2}}, \text{shift}=+1, \text{dim}=d), \\
    &X_{:,:,:,\frac{F}{2}:,} 
\end{aligned}
\right)
\end{equation}

\begin{itemize}
    \item \textbf{Intra-Bin Phase} ($d=2$): Processes temporal relationships within bins
    \begin{equation}
    X_{intra} = \mathcal{T}(X, 2) \in \mathbb{R}^{Bt \times T \times Bn \times F},
    \end{equation}
    
    \item \textbf{Inter-Bin Phase} ($d=1$): Captures cross-bin dependencies
    \begin{equation}
    X_{final} = \mathcal{T}(X_{intra}, 1) \in \mathbb{R}^{Bt \times T \times Bn \times F},
    \end{equation}
\end{itemize}

\noindent Unified components:
\begin{itemize}
    \item $\text{roll}(\cdot)$: Cyclic shift along specified axis (negative=left, positive=right).
    \item Channel split: First $F/4$ left-shifted, next $F/4$ right-shifted, last $F/2$ static.
    \item Dimensions: $Bt$=batch size, $T$=bin count, $Bn$=frames/bin, $F$=features.
\end{itemize}

\subsubsection{Mechanism Analysis}
The dual-shift design establishes:
\begin{itemize}
    \item \textbf{Local Connectivity}: The intra-bin shift operation captures short-term motion patterns with fine temporal granularity, ensuring that local temporal dynamics are effectively modeled.
    \item \textbf{Global Context}: The inter-bin shift operation propagates features across distant segments, enabling the capture of long-range temporal dependencies and holistic temporal understanding.
    \item \textbf{Channel-Specific Dynamics}: 50\% channels ($\frac{F}{2}$) remain unshifted to preserve spatial semantics, ensuring that the spatial context is not lost during the temporal modeling process.
\end{itemize}
This design choice allows the BTSM to reconcile the fine-grained motion dynamics captured by event sensing with the need for holistic temporal reasoning, making it particularly effective for event-based gesture recognition tasks. The module's parameter-free nature and lightweight computation make it a practical solution for real-world applications.

\subsection{Loss Function}
We employ standard cross-entropy loss for gesture classification:
\begin{equation}
\mathcal{L} = -\sum_{c=1}^C y_c \log(p_c),
\end{equation}
where \(y_c\) denotes ground-truth labels and \(p_c\) represents predicted class probabilities.

\section{Experiments}
\subsection{Dataset Configuration and Evaluation Protocol}
Our data splitting methodology focuses on two distinct evaluation dimensions: illumination robustness evaluation and two different training-test split methods, the heterogeneous split and the homogeneous split.

\subsubsection{Illumination Robustness Validation}
We validate illumination robustness through three test scenarios:
\begin{itemize}
  \item \textbf{Normal light}: Baseline performance evaluation.
  \item \textbf{Low light}: Assess how the model handles low light conditions.
  \item \textbf{Mixed}: Real-world simulation with combined conditions.
\end{itemize}

\subsubsection{Training-Test Split Protocols}
Two evaluation paradigms implemented:
\begin{itemize}
  \item \textbf{Heterogeneous Split (Generalization)}: 
  \begin{itemize}
    \item Test set: Subjects 1,2,9 (unseen identities).
    \item Trains on remaining 7 subjects.
    \item Evaluates cross-subject robustness and the model's generalization ability across different identities and lighting environments.
  \end{itemize}
  
  \item \textbf{Homogeneous Split (Adaptability)}:
  \begin{itemize}
    \item 3:1 split per subject's sequences.
    \item Maintains identity consistency and validates method effectiveness.
    \item Tests illumination variation handling.
  \end{itemize}
\end{itemize}

\subsection{Main Results on Proposed Dataset}
\begin{figure}
  \centering
  \includegraphics[width=0.98\linewidth]{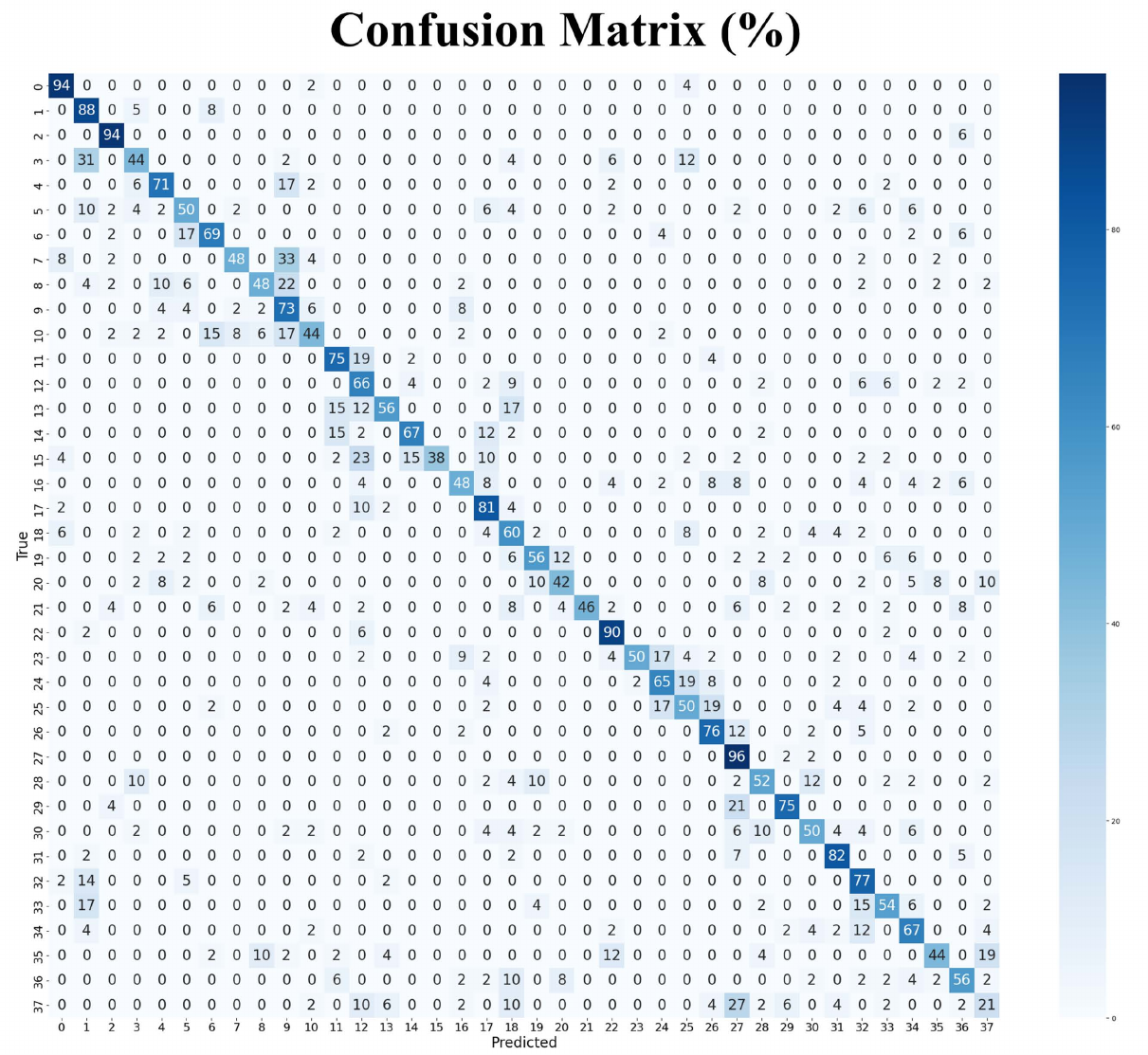}
  \caption{Confusion matrix for mixed-condition testing in Heter split using our method. Diagonal entries represent recall rates for 38 activity classes, while off-diagonal values show common misclassifications.}
  \label{fig:conf_matrix}
  \vspace{-1.5em}
\end{figure}

Figure~\ref{fig:conf_matrix} presents the normalized confusion matrix for our proposed method testing in the Heter split. 
The diagonal elements represent correct classifications, whereas off-diagonal entries indicate misclassifications. Our model demonstrates strong discriminative capabilities, particularly in wavings and numbers, achieving classification accuracies above 90\%. 
Notable misclassifications occur primarily in freestyle motions, likely attributed to high inter-personal variability in movement execution and the test set containing completely unseen movement patterns that differ significantly from training data distributions. Other confusion occurs mainly between visually similar activities like Push forward (class 15) and Wave up (class 12) or Raise arms (class 14).
All experiments are conducted on a single A100 GPU using the AdamW optimizer with a learning rate of \(1 \times 10^{-4}\), trained for 60 epochs with early stopping to prevent overfitting.

\subsection{Comparative Methods and Results}

We conduct comprehensive comparisons with four approaches that address distinct aspects of vision-based action recognition particularly relevant to action recognition under varying conditions.

\begin{itemize}
\item \textbf{S-tllr}~\cite{apolinario2023s}: A spiking neural network (SNN) designed for event-based action recognition. This biologically inspired approach achieves state-of-the-art performance on the DVS128 dataset.
  
\item \textbf{Assembly}~\cite{ohkawa2023assemblyhands}: A strong baseline derived from the AssemblyHands dataset, originally designed for RGB-based action recognition using ResNet-101 backbone.
  
\item \textbf{EventEgo3D}~\cite{millerdurai2024eventego3d}: The first framework addressing egocentric 3D perception tasks with event cameras, combining LNES event representation with efficient BlazeBlocks for spatial-temporal feature extraction. Given its strong alignment with our task requirements, we select it as our baseline method.

\item \textbf{EventPoint}~\cite{chen2022efficient}: A point cloud approach for event data processing that achieves competitive performance on the DHP19 dataset with low computational overhead.
\end{itemize}

\begin{table}[!t]
\centering
\caption{Cross-method performance comparison (Accuracy \%).}
\label{tab:main_results}
\scriptsize 
\setlength{\tabcolsep}{3pt} 
\renewcommand{\arraystretch}{1.1} 

\begin{tabular}{@{}l
                >{\centering\arraybackslash}m{1.1cm}
                >{\centering\arraybackslash}m{1.4cm}
                >{\centering\arraybackslash}m{1.1cm}
                >{\centering\arraybackslash}m{1.4cm}
                >{\centering\arraybackslash}m{0.9cm}@{}}
\toprule
\multirow{2}{*}{Method} & 
\multicolumn{2}{c}{Heter Split} & 
\multicolumn{2}{c}{Homo Split} & 
\multirow{2}{*}{Params} \\
\cmidrule(lr){2-3} \cmidrule(lr){4-5}
 & Mixed & Day/Night & Mixed & Day/Night & (M) \\
\midrule
S-tllr~\cite{apolinario2023s} & 30.2 & 28.4/27.9 & 50.0 & 49.7/50.1 & 13.96 \\
Assembly~\cite{ohkawa2023assemblyhands} & 59.6 & 59.8/58.6 & 98.9 & 98.9/98.9 & 42.72 \\
EventEgo3D~\cite{millerdurai2024eventego3d} & 36.4 & 39.5/38.3 & 97.4 & 97.8/97.2 & 6.50 \\
EventPoint~\cite{chen2022efficient} & 30.0 & 29.3/28.1 & 98.6 & 98.4/98.7 & 3.35 \\
\midrule
Ours & \textbf{62.7} & \textbf{62.8/62.5} & \textbf{99.1} & \textbf{99.2/98.9} & 7.01 \\
\bottomrule
\end{tabular}

\vspace{0.3em}
\footnotesize 
\begin{tabular}{@{}p{0.95\linewidth}@{}}
*\,\textit{EventEgo3D}: Adapted from 3D keypoint estimation (original feature extractor + our classification head). \\
\end{tabular}
\vspace{-1.5em}
\end{table}

Table \ref{tab:main_results} presents the performance comparison across different methods. Our method achieves the best performance in both heterogeneous and homogeneous splits. For the heterogeneous Split, our method achieves 62.7\% accuracy on the Mixed-condition test, significantly outperforming other methods such as STLLer (30.2\%), Assembly (59.6\%), EventEgo3D (36.4\%), and EventPoint (30.0\%). Consistent normal light/ low light performance (62.8\% day vs 62.5\% night) demonstrates illumination robustness. All methods except S-tllr achieve an accuracy higher than 97\% in the homo split validating experimental consistency. The parameter count of our model is 7.01M, which is relatively small, indicating that our model is efficient and effective for event-based egocentric action gesture recognition. 
In conclusion, our method outperforms existing approaches in action recognition under varying conditions while maintaining a relatively small number of parameters, demonstrating its robustness and effectiveness.

\subsection{Ablation Study}
As shown in Table \ref{tab:ablation}, our ablation study reveals the effectiveness of different modules. The baseline model achieves 36.4\% accuracy on the Heterogeneous Mixed testing set. Introducing the BTSM boosts the accuracy to 58.1\%, demonstrating the importance of temporal information in action recognition. Adding the Visual Mamba module alone improves the accuracy to 45.3\%, indicating the value of spatial feature extraction. The full model combining BTSM and VMamba achieves the highest accuracy of 62.7\%, showing that both modules contribute significantly to performance enhancement.

\begin{table}[!h]
\centering
\caption{Module contribution in heterogeneous mixed testing (Accuracy\%).}
\label{tab:ablation}
\begin{tabular}{l c}
\toprule
Configuration & Mixed Accuracy \\
\midrule
Baseline & 36.4 \\
+BTSM only & 58.1 (+21.7) \\
+VMamba only & 45.3 (+8.9) \\
Full Model & \textbf{62.7 (+26.3)} \\
\bottomrule
\end{tabular}
\vspace{-1.5em}
\end{table}

\subsection{Quantitative Result on DVS128 Gesture}
To validate the generalization capability of our proposed architecture beyond our proprietary dataset, we conduct extensive experiments on the widely recognized event-based gesture recognition benchmark, DVS128 Gesture~\cite{amir2017low}. This dataset consists of 1,342 instances of 11 distinct hand gestures captured using dynamic vision sensors, presenting unique challenges in sparse temporal feature learning.
As shown in Table~\ref{tab:results}, our approach achieves a remarkable accuracy of 97.0\% on this benchmark, demonstrating its strong generalization capability. This result matches the accuracy of 97.0\% obtained through reimplementation of the open source state-of-the-art approach S-tllr~\cite{apolinario2023s} without access to its pre-trained models, despite their reported accuracy of 97.7\%. 
For S-tllr, we follow the settings described in its official implementation. For all other methods, experiments are conducted on a single A100 GPU using the AdamW optimizer with a learning rate of \(1 \times 10^{-4}\), trained for 60 epochs with early stopping to prevent overfitting.

\begin{table}[t!]
\centering
\caption{Performance comparison on the DVS128 Gesture benchmark. Our method achieves state-of-the-art accuracy without pretrained models, comparable to the reimplemented state-of-the-art S-tllr and other methods. Results highlight robustness to temporal sparsity and gesture variability.\\
* denotes our reimplementation.}
\label{tab:results}

\begin{tabular}{lc}
\toprule
\textbf{Method} & \textbf{Accuracy (\%)} \\
\midrule
Assembly~\cite{ohkawa2023assemblyhands} & 91.3 \\
EventEgo3D~\cite{millerdurai2024eventego3d} & 81.8 \\
EventPoint~\cite{chen2022efficient} & 76.1 \\
S-tllr~\cite{apolinario2023s}* & 97.0 \\
Ours & 97.0 \\
\bottomrule
\end{tabular}

\vspace{-1.5em}
\end{table}

Our leading recognition performance on the DVS128 Gesture dataset stems from three key architectural innovations: (1) A lightweight CNN with asymmetric depthwise convolutions that maintains spatiotemporal features while reducing parameters, crucial for sparse event-based data; (2) A plug-and-play state-space context block that focuses on gesture dynamics; (3) The parameter-free BTSM module that effectively fuses sparse spatiotemporal events. Achieving 97.0\% accuracy (matching S-tllr's reimplemented performance) validates our architecture's generalization capability and robustness for event-based gesture recognition.

\section{Conclusion}
We propose a novel solution for event-based egocentric gesture recognition, addressing its key challenges through three technical contributions.
In this work, we address the new problem, i.e., event-based egocentric gesture recognition based on egocentric event cameras by proposing an egocentric gesture recognition framework featuring a three-stage processing pipeline: First, raw event streams are converted into Locally Normalized Event Surfaces (LNES) to suppress ego-motion artifacts from head movements. Subsequently, a lightweight Blaze encoder-decoder combined with a VMamba module extracts spatiotemporal features, addressing event sparsity and long-term temporal modeling. Finally, a parameter-free BTSM achieves motion-invariant temporal alignment through bidirectional propagation. To address the data gap, we construct EgoEvGesture, the first large-scale egocentric event-based gesture dataset, containing 5,419 samples of 38 gesture classes (single/two-hand interactions) from 10 subjects under normal and low light conditions. Experiments demonstrate that our method achieves 62.7\% accuracy in heterogeneous testing with only 7M parameters, outperforming state-of-the-art approaches by 3.1\%, while maintaining stable performance. 
As this is the first work on egocentric gesture recognition using event cameras, the open-source release of the dataset and code is expected to foster synergistic advancements in event vision and egocentric interaction research. 

In the future, we intend to enhance the head-mounted capture system to make it more lightweight. 
We plan to leverage depth cameras as ground truth to explore egocentric hand keypoint estimation based on event cameras, and then extend the relevant findings to outdoor environments.

\bibliographystyle{IEEEtran}
\bibliography{paper}

\clearpage
\section{Comprehensive Event Rate Analysis for EgoEvGesture Dataset}

The dataset is characterized by its high variability as shown in the following figures, which uses different normalized time event rates to reveals three fundamental characteristics through vertically stacked visualizations, including different people performing different actions event rates, different people's average event rates across all actions in different environments, and event rates for unilateral and bilateral actions.

\begin{figure*}[h]
  \centering
  \includegraphics[width=0.85\linewidth]{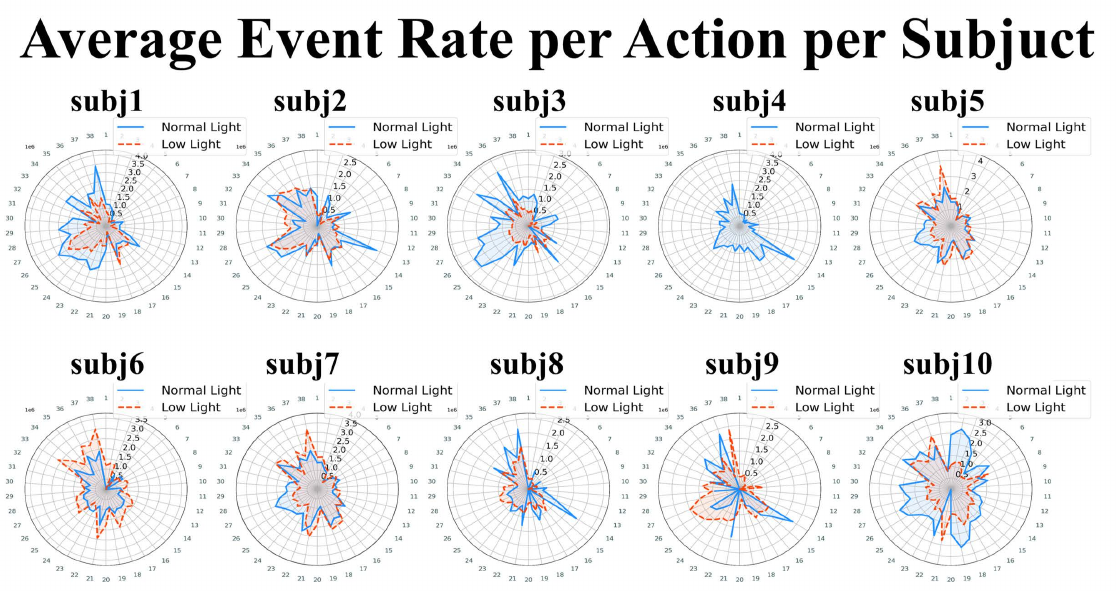}
  \caption{Average action event rates under normal light and low light conditions across 10 subjects. Each subplot represents data from a single subject, with 38 discrete actions uniformly distributed around the 360{\textdegree} axis. The 10$\times$ difference in event rate validates both inter-subject variability and intra-subject variability.}
  \label{fig:action_specificity_supp}
\end{figure*}

Figure~\ref{fig:action_specificity_supp} shows substantial differences in normalized event rates both when different subjects perform the same action (inter-subject variability) and when individual subjects perform different actions (intra-subject variability), with up to 10$\times$ variation between simplest and most complex gestures, demonstrating the distinctiveness of the action design.

\begin{figure*}[h]
  \centering
  \includegraphics[width=0.85\linewidth]{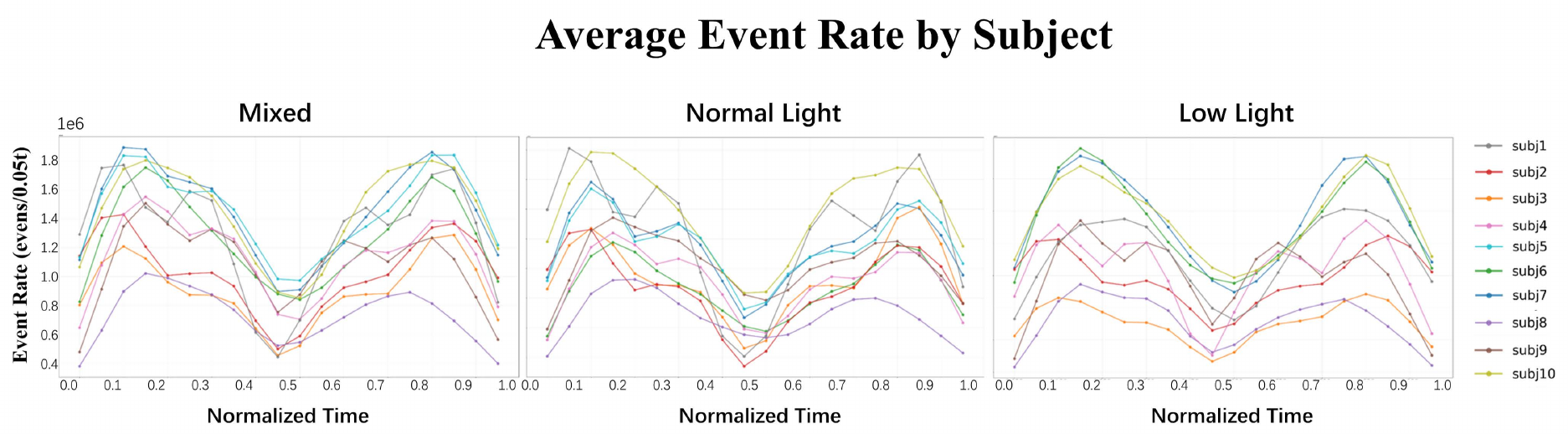}
  \caption{Subject-wise event rates under mixed, normal light, and low light conditions showing individual adaptation variance.}
  \label{fig:subject_variability}
\end{figure*}

Figure~\ref{fig:subject_variability} shows a clear difference in the average event rates between different subjects, regardless of whether it is normal light or low light.

\begin{figure*}[h]
  \centering
  \includegraphics[width=0.85\linewidth]{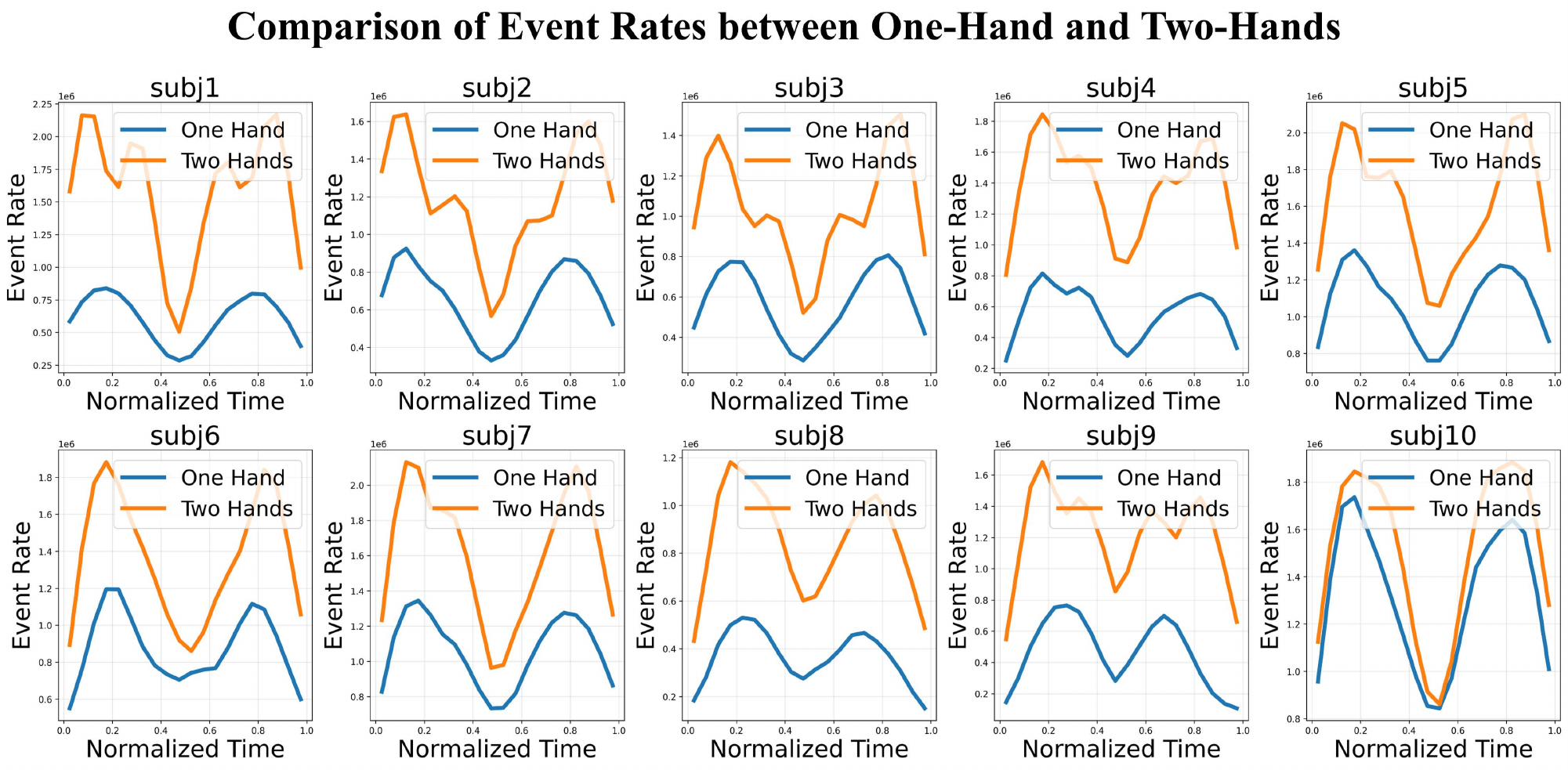}
  \caption{Bimanual vs unimanual event rates. Higher event density in two-hand gestures confirms asymmetric motion dynamics.}
  \label{fig:hand_assymetry}
\end{figure*}

Figure~\ref{fig:hand_assymetry} shows that the event rate for bilateral actions is significantly higher than that for unilateral actions, demonstrating the difference between the two types of actions. This variability is crucial for training robust gesture recognition models that can be generalized to different users and lighting conditions. The inclusion of event camera data, which is less sensitive to lighting variations compared to traditional RGB cameras, further enhances the dataset's utility for research in challenging environments.

\end{document}